# Question Identification in Arabic Language Using Emotional Based Features

Ahmed Ramzy*[1], and Ahmed Elazab[2]
[1]Cairo University, Giza, Egypt
a.ramzy@fci-cu.edu.eg
[2]Fingerprint Consultancy, Cairo, Egypt
aelazab@fpconsultancy.com

**Abstract**. With the growth of content on social media networks, enterprises and services providers have become interested in identifying the questions of their customers. Tracking these questions become very challenging with the growth of text that grows directly proportional to the increase of Arabic users thus making it very difficult to be tracked manually. By automatic identifying the questions seeking answers on the social media networks and defining their category, we can automatically answer them by finding an existing answer or even routing them to those responsible for answering those questions in the customer service. This will result in saving the time and the effort and enhancing the customer feedback and improving the business. In this paper, we have implemented a binary classifier to classify Arabic text to either question seeking answer or not. We have added emotional based features to the state of the art features. Experimental evaluation has done and showed that these emotional features have improved the accuracy of the classifier.

## 1 Introduction

Several research studies have shown that most of the social media users are increasingly writing text containing questions [12, 6]. However, not all of them are seeking answers [14]. Identifying and tracking these questions seeking answers is very difficult task in case that a person has to answer this large set of questions.

We aim to identify the Arabic text that contains question seeking answers. There may be a text that contains question keyword but not seeking answers such as advertisements, quoted questions or questions with answers. Consequently, we need to identify the questions that seeking answers. We can provide guidance for decision makers to know the most areas that the users are asking about them and introduce solutions and improvements. As well, we can route the questions only that are seeking answers to the customer support team, this will result in saving the time and the efforts. Therefore enhancing the customer feedback.

In this paper, we proposed a machine learning approach to analyze the Arabic text of the social media content (tweets, posts, and comments). We built a binary classifier to identify whether the comment, post, or tweet is question seeking answer or not. We have trained this classifier using lexical features, input text related features and emotional based features. Adding the emotional based features is a novel approach in the questions identification research. To evaluate and show the effect of adding this category of features, we implement the classifier with the state of the art features without adding the emotional features. We implemented the model using more than one type of classifiers such as SVC, LinearSVC, Naïve Bayes, LogisticRegression, and GaussianNB. We compared our results before and after using the emotional features. We have tested the model classifiers on more than one datasets from different sectors. The results showed that adding the emotional features increase the accuracy of the model regardless the type or the efficiency of the classifier as shown in Table 2.

The main contribution of this paper is the using of emotional based features to improve the accuracy of the binary classifier regardless the type of the classifier. The rest of this paper is organized as follows. Related work is discussed in Section 2. The detailed key contributions of the paper are presented in sections 3 where we describe our methodology. Section 4 we discuss the experimental evaluation. Finally section 5 concludes the paper with a critical discussion and an outlook on future work.

## 2 Literature review

The problem of identifying the questions that seek answers over the English text in User Generated Content (UGC) has been studied in several researches. Either by using rule based approaches in Efron M et al [7], or machine learning approaches in [5, 13, 2]. As well, there are some researcher used hybrid approach like in Arun D P et al [2]. They have employed the traditional rule based approach and learning based method that constructs a binary classifier with lexical and syntactic features including question, context, and tweet-specific features. Recently, Hasanain M et al [8] attempted the first study on Arabic tweets to identify the questions that seeking answers. They first used a rule-based filter to extract tweets with question keywords. Then they developed a binary classifier to detect tweets with answer-seeking questions. They trained this classifier using set of features including lexical, structural, question-specific, tweet-specific, and (in) formality aspects of the tweets. They have achieved 0.64 in recall, 0.80 in precision and 0.72 in F-measure. Likewise, there is a demo paper [1] has been proposed to identify the questions from twitter data streams. Question identification and classification research has been used also in the Academic and the E-learning forums [11].

Some researchers like Zhao Z et al [14] are interested in understanding the questions of web users. By investigating research questions about these questions like "How are they distinctive from daily posted conversation?", "How are they related to search queries?" They developed automated text classifier so they can accurately detect the real questions in tweets. Finally, the researches of finding the semantic categories of the questions in English were proposed in several papers [3, 10].

## 3 Methodology

In this section, we present a model for implementing question identification by augmenting a machine learning approach with a set of features derived from Arabic text lexicon and input text based features as well as emotional based features.

### 3.1 Model component

We employed the statistical machine learning for identifying the text that contains questions seeking answers. From the previous work presented in [9, 8, 14] that discussing the learning approach, it was revealed that classifiers that perform well for the task of question identification are the family of Naïve Bayes as well as SVM. In our model, we have chosen to use Naïve Bayes classifier based on experiments that are comparing the accuracy of the two classifiers to show which the best classifier is top performing across different datasets. We carried out the system using python sklearn library [4].

The first step in our model is the text preprocessing followed by features extraction process. These features inputted to Naïve Bayes classifier. We have used three types of features:

a) The lexical features such as unigrams, bigrams and trigrams.
b) Input text related features like the proposed work of Hasanain M et al [8], such as the text length, number of questions marks and exclamation marks, the count of emojis and emoticons (such as: :-)) to measure the formality of the text, etc.
c) One of our main contributions is using the emotional based features to improve the accuracy of the Naïve Bayes classifier such as the occurrence of sentiment words or phrases within the input text.

We will explain each of the above features later in section 3.3 in details. The following subsection, we present the preprocessing steps that were carried out in our model.

### 3.2 Preprocessing steps

In this subsection, we present a detailed description of the preprocessing procedures for our question identification classifier. The preprocessing process consists of more than one procedure.

a) Text cleaning: Here, we are removing the special characters, the punctuations, non-Arabic characters, non-Unicode characters, etc.
b) Stop words removing: Here, we remove the words which don't add any meaning to the classifier, such as "على" ,"في" and "من" , like the words "in", "on" and "from" in English.
c) Text normalization: In this step, the diacritic are removed, also we replaced the characters "إ" ,"أ" and "آ" with "ا". The characters "ئ" and "ؤ" with "ء". As well as the character "ي" is replaced with "ى" and the character "ة" is replaced with "ه".
d) Elongation removal: In social media, the elongation is a way to express a word with more emphasis. For example "لطيييييييييييف", like the word "niiiiiiiiiiiice" in English. We applied an algorithm to detect if there are three or more consecutive repeated character. If so, the consecutive repetitions are replaced by a single instance of that character. So, the word "niiiiiiiiiiiice" will be transformed to "nice" and "لطيييييييييييف" to "لطيف".

### 3.3 Feature extraction

We used some features which can be categorized in three categories:

1. Lexical Features

   Here, we used unigrams, bigrams and trigrams represented with their IDF weights. Counting on the unigrams, bigrams and the trigrams instead of using only the unigrams to add more semantics to the classification algorithm.

2. Input text related features

   The following are list of the input text related features used by our model. We've used some of the input related features such as the length of the text in characters, the length of the text in words, the count of the segments within the input text. We are assuming that the segment are separated by one or more of the following characters "-?!.؟: etc.". Number which set to 1 if the input text contains URL and to 0 otherwise [9, 14], number which set to 1 if the input text starts with URL and to a 0 otherwise, number which set to 1 if the input text ends with URL and to a 0 otherwise, number indicates the count of question keywords hashtags like #سؤال like #question in English [8], Feature indicates the count of mentions in the text, Feature indicates the count of hashtags in the text, Feature indicates the count of the quoted strings in input text [8], Number indicates the count of diacritics in the input text, number indicates the count of the question keywords included in the input text for example "كيف". We have used the collected question phrases from [8].

3. Emotional based features

   As we mentioned before, one of our contributions is that we used emotional based features such as the occurrence of sentiment words or phrases within the input text. We found that this type of features affect in improving the accuracy of the Naïve Bayes classifier. Based on our analysis of the results, we noticed that the text that is containing emotional features may be written to express an opinion and not written to ask about particular question. We have manually built Arabic sentiment lexicon which has positive and negative entries including sentiments terms, compound_terms, emojis and emoticons. The following are list of the emotional based features used by our model.

- numOfPos: feature indicates the number of positive terms or compound terms within the input text that have matched with the positive entries in the sentiment lexicon.
- numOfNeg: features indicates the number of negative terms or compound_terms within the input text that matched with the negative entries in the sentiment lexicon.
- startWithPos: Number which set to 1 if the input text starts with positive terms or compound terms within the input text that have matched with the positive entries in the sentiment lexicon and to a 0 otherwise.
- startWithNeg: Number which set to 1 if the input text starts with negative terms or compound terms within the input text that have matched with the negative entries in the sentiment lexicon and to a 0 otherwise.
- endWithPos: Number which set to 1 if the input text ends with positive terms or compound terms within the input text that have matched with the positive entries in the sentiment lexicon and to a 0 otherwise.
- endWithNeg: Number which set to 1 if the input text ends with negative terms or compound terms within the input text that have matched with the negative entries in the sentiment lexicon and to a 0 otherwise.
- posPercentage: Number represents the percentage of terms in the text that are positive. We normalize this number from 0 to 1.
- negPercentage: Number represents the percentage of terms in the input text that are negative. We normalize this number from 0 to 1.
- numOfPosEmo: Number indicates the count of positive emojis and emoticons within the input text that matched with the positive emojis and emoticons entries in the sentiment lexicon. The features used in [8] includes the number of emojios and emoticons within the text regardless they are positive or negative. The point behind that is to measure the level of formality in the input text. In our work we count the positive ones as well as the negative ones.
- numOfNegEmo: Number indicates the count of negative emojis and emoticons within the input text.

## 4 Experiments and Results

The goal of the presented work in this paper is to determine whether the emotional features improved the question identification model results or not. To determine this, we implemented the model using the features used in the state of the art specifically in Hasanain M et al [8]. We've considered the previous step as the baseline. After that we've added the emotional features and compared it by the model implemented without emotional features. We've used the dataset from Hasanain M et al [8]. This dataset is considered as generic tweets. Also we've collected other datasets from different sector like banking, telecom and retail sectors. We are describing these used datasets in Table 1. Note that our experiments were implemented using the python scikit-learn library [4].

**Table 1.** Datasets used in our experiment.

| Sector | Source | No. Of Records |
|---|---|---|
| Generic | Twitter | 5000 |
| Banking | Twitter & Facebook | 2500 |
| Telecom | Twitter & Facebook | 1500 |
| Retail | Facebook | 1000 |

We have manually annotated the collected datasets. We generally divide our datasets into training and testing sets according to 80/20 rule. We tested our model which relies on the features presented in section 3.3. The results of our model are presented in different convolution matrix shown in Table 2. Note that P, R and F refer to Precision, Recall and F-measure respectively.

**Table 2.** Comparason between the acurracy before and after adding the emotional features.

| | | Models | | | | | | | | | |
|---|---|---|---|---|---|---|---|---|---|---|---|
| | | SVM | | Linear SVM | | Naive Bayes | | Logistic Regression | | GaussianNB | |
| Datasets | | Before | after | before | after | Before | after | before | after | before | After |
| Generic | P | 0.76 | 0.79 | 0.82 | 0.84 | 0.89 | 0.93 | 0.88 | 0.89 | 0.71 | 0.73 |
| | R | 0.70 | 0.73 | 0.68 | 0.73 | 0.90 | 0.93 | 0.59 | 0.60 | 0.64 | 0.65 |
| | F | 0.73 | 0.76 | 0.74 | 0.78 | 0.89 | 0.93 | 0.71 | 0.72 | 0.67 | 0.69 |
| Banking | P | 0.82 | 0.86 | 0.85 | 0.87 | 0.84 | 0.88 | 0.86 | 0.87 | 0.78 | 0.79 |
| | R | 0.81 | 0.86 | 0.85 | 0.87 | 0.83 | 0.86 | 0.85 | 0.86 | 0.78 | 0.79 |
| | F | 0.81 | 0.86 | 0.85 | 0.87 | 0.83 | 0.87 | 0.85 | 0.86 | 0.78 | 0.79 |
| Telecom | P | 0.70 | 0.74 | 0.69 | 0.74 | 0.71 | 0.75 | 0.71 | 0.73 | 0.62 | 0.63 |
| | R | 0.68 | 0.73 | 0.67 | 0.74 | 0.72 | 0.75 | 0.66 | 0.67 | 0.61 | 0.62 |
| | F | 0.69 | 0.73 | 0.68 | 0.74 | 0.71 | 0.75 | 0.68 | 0.70 | 0.61 | 0.62 |
| Retail | P | 0.67 | 0.73 | 0.71 | 0.75 | 0.82 | 0.85 | 0.74 | 0.75 | 0.59 | 0.61 |
| | R | 0.68 | 0.74 | 0.72 | 0.75 | 0.79 | 0.82 | 0.71 | 0.72 | 0.59 | 0.61 |
| | F | 0.67 | 0.73 | 0.71 | 0.75 | 0.80 | 0.83 | 0.72 | 0.73 | 0.59 | 0.61 |

## 5 Conclusion and Future Work

In this paper we presented question identification classifier that can be used to identify the questions seeking answers. We have presented a set of features which included emotional features in order to enhance the questions identification model. Adding emotional features is a novel approach in question identification classifiers. We believe that adding the emotional features had an impact in enhancing the results. We have shown that when adding the emotional features, the accuracy of the model increased. We have tested the model on a benchmark dataset from Hasanain M et al [8]. Also another different datasets from different sectors like banking, telecom and retail sectors.

In the future we would like to perform some analytics on the identified questions text to know the entities and the aspects which the question is asking about them. Also, we would like to implement a semantic categorization of the text that contains question seeking answers. By automatic identifying the questions seeking answers on the social media networks and defining their category, we can automatically answer them by finding an existing answer in a knowledge base or even routing them to those responsible for answering those questions in the customer service.

## Acknowledgment

This work is funded by R&D department in Fingerprint Consultancy Company.